\documentclass[12pt]{article}

\usepackage{graphicx} 
\usepackage{authblk}
\usepackage{amsthm}
\usepackage{amsmath}
\usepackage{algorithm}  
\usepackage{algorithmic}

\newtheorem{definition}{\textbf{Definition}}

\newtheorem{theorem}{\textbf{Theorem}}

\newtheorem{corollary}{\textbf{Corollary}}
\newtheorem{remark}{\textbf{Remark}}

\title{Dependence Structure Estimation via Copula}

\author[1]{Ma Jian\thanks{Email: majian@hitachi.cn.}}
\author[2]{Sun Zengqi}
\affil[1]{\normalsize Hitachi (China) Research \& Development Corporation}
\affil[2]{\normalsize Tsinghua University}

\begin{document}

\maketitle

\begin{abstract}
\noindent
Dependence strucuture estimation is one of the important problems in machine learning domain and has many applications in different scientific areas. In this paper, a theoretical framework for such estimation based on copula and copula entropy -- the probabilistic theory of representation and measurement of statistical dependence, is proposed. Graphical models are considered as a special case of the copula framework. A method of the framework for estimating maximum spanning copula is proposed. Due to copula, the method is irrelevant to the properties of individual variables, insensitive to outlier and able to deal with non-Gaussianity. Experiments on both simulated data and real dataset demonstrated the effectiveness of the proposed method. 
\end{abstract}

\section{Introduction}
Dependence between random variables is of significant importance since it may means essential statistical or causal relationships within real world social, physical, or biological systems. As data are collected from different scientific fields, analyzing and understanding them remains a challenge. Hence, dependence structure learning is becoming one of the most important problems in machine learning domain.

The most studied statistical methodology for dependence representation has been graphical models, or Bayesian networks \cite{1,2,3}. With graphical models formalism, a probability density is represented with a directed or indirected graph, of which each node represents a random variable, and each edge represents a conditional dependence relation between two random variables.  Such representation lead to simplicity of models and hence makes large-scale problem modeling and inferring tractable. The assumption of graphical models is markovity or conditional independence, which means only first order dependence or pairwise dependence is considered in those models. However, the assumption may be incorrect in many real applications.

Additionally, traditional learning methods for inferring graphical models involve maximum likelihood where one should specify parametric distribution family with parametric margins of individual variables implicitly. Model selection on margins is critical for structure learning to a large extent, but there is usually short of priori knowledge needed for such selection. So we are interested in finding a method that can separate structure learning from parametric marginal specification.

Copula theory is on the representation of multivariate statistical dependence \cite{joe1997,nelsen1998}. According to Sklar theorem \cite{6}, any multivariate probilistic distribution can be represented as a product of its margins and a copula function which represents dependence structure among random variables. With copula, one can separate the margins from their joint density distribution and dependence structures, and therefore it is possible to study only statistical relationships without considering individual properties of each variable. Copula has been widely studied in finance \cite{bouye2000}, and gain momentum in machine learning community \cite{ma2007,kirshner2008}. Since copula is a unified theory on representation of statistical dependence, it is natural to build a universal framework for structure learning with it. Apparently, copula representation covers all the representations of statistical dependence, including graphical models.

Copula Entropy (CE) is a recently introduced theory on statistical independence measurement \cite{ma2011}. It enjoys many properties which an ideal dependence measure should have, including multivariate, symmetric, non-negative (0 iff independent), invariant to monotonitic transformation, equivalent to correlation coefficient in Gaussian cases. Ma and Sun proved its equivalence to Mutual Information (MI) and then proposed a simple and elegant estimation method \cite{ma2011}. With CE, one can infer statistical dependence relationships between random variables without bothering marginal properties.

The main contribution of the paper is introducing a novel framework of structure learning based on copula and copula entropy. Such framework provides a universal theory of structure learning. In the paper, graphical models is identified as a special case of the framework, while the copula framework covers all types of dependence structure theoretically. Particularly, graphical models concerns only pairwise dependence, and has its counterpart in copula theory, called product copula. 

We study estimating dependence structure with CE. We propose a method in which dependence structure is first measured with CE and then infer maximum spanning copula with Chow-Liu \cite{11} like MST algorithms on dependence matrix. The most advantage of CE is that it is a model-free measure with non-parametric estimation. Moreover, our methods can be generalized to estimate much complex dependence relationships.  

\section{Copula Theory}

\subsection{Definitions and Theorem}
Copulas are the functions that model the dependence relations among random variables, and is defined as follows:

\begin{definition}[Copula] \cite{joe1997,nelsen1998}
	Given $N$ random variables $\mathbf{X}=\{X_1,\ldots,X_N\} \in \mathcal{R}^N$. Let $\{u_i=F_i(x_i),i=1,\ldots,N\}$ be the marginal distributions of $\mathbf{X}$. A $N$-dimensional copula $C: \mathcal{I}^N\rightarrow \mathcal{I}$ ($\mathcal{I}=[0,1]$) of $\mathbf{X}$ is a function with following properties:
	\begin{itemize}
		\item $C$ is \textit{grounded} and \textit{N-increasing};
		\item $C(1,\ldots,1,u_i,1,\ldots,1)=u_i$.
	\end{itemize}
\end{definition}

Intuitively, copula can be viewed as a cumulative distribution function (CDF) stretched onto $\mathbf{u}=\mathcal{I}^N$ from the CDF of $\mathbf{X}$. 

The relation between CDF, margins, and copula is stated in Sklar's theorem \cite{6}:

\begin{theorem}[Sklar's Theorem]
	Given a random vector $\mathbf{X}=\{X_1,\ldots,X_N\}$, its CDF $\mathbf{F}(\mathbf{x})$ can be represented as
	\begin{equation}
	\mathbf{F}(\mathbf{x})=C(u_1,\ldots,u_N),
	\label{eq:sklarthm}
	\end{equation}
	where $C$ is a copula function, $\{u_i\}$ are marginal distribution functions of $\mathbf{X}$. If $\{F_i\}$ are continuous, then $C$ is unique.
\end{theorem}

Sklar's theorem is of central importance in copula theory. By applying derivative on (\ref{eq:sklarthm}), one can also represent probability density function (PDF) via copula. Let us first define \textit{copula density}.

\begin{definition}[Copula Density]
	A $N$ dimensional \textit{copula density} $\mathbf{c}$ corresponding to $N$-copula $\mathbf{C}$ is defined as
	\begin{equation}
	c(\mathbf{u})=\frac{d^N}{d{u_1},\ldots,d{u_N}}C(\mathbf{u}), 
	\end{equation}
	where $\mathbf{u} \in \mathcal{I}^N$.
	\label{eq:copuladensity}
\end{definition}

With copula density, one can derive a corollary of Sklar's theorem:
\begin{corollary}
	The probability density function (PDF) $p(x)$ of $\mathbf{X}$ can be represented as:
	\begin{equation}
	p(\mathbf{x})=\mathbf{c}(\mathbf{u})\prod_{i=1}^{N}{p_i(x_i)}
	\label{eq:pdfcopula}
	\end{equation}
	where $\{p_i,i=1,\ldots,N\}$ are marginal density functions of $\mathbf{X}$, and $c$ is copula density.
\end{corollary}

\begin{remark}
	According to Sklar's theorem, dependence structure is independent from margins. This implies that it is possible that a same structure is associated with different distributions. These distributions are said to be \textit{equivalent} in a sense of copula. That means that different distribution may corresponds to same dependence structure.
\end{remark}


\subsection{Product Copula}
As dependence representation, there are many types of copula function and the methods for contructing copulas \cite{nelsen1998}. Here we introduce a special type of copula, product copula.

\label{sec:ProductCopula}
\begin{theorem}[Product Copula]
	The product of copula density of independent variables is also a copula density.
\end{theorem}
\noindent
The theorem can be represented as
\begin{equation}
c(\mathbf{u})=\prod_{m=1}^{M}{c_m(\mathbf{u}_m)}.
\label{eq:prodcop}
\end{equation}
where $\{c_m\}$ are any type of copula density, and $\mathbf{u}=\cup_{m=1}^{M}{\mathbf{u}_m}$, and $\{\mathbf{u}_m\}$ are vectors of margins of random variables. If all the sub-copulas $\mathbf{c}_m$ are bivariate, it means that there is only pairwise dependence exists. In this case, product copula is equivalent to a graphical model.
\begin{theorem}
	Any graphical model is equivalent to a product copula with only bivariate sub-copulas.
	\label{thm:gmprodcop}
\end{theorem}
\noindent
The theorem indicates that graphical model is just a special case of copula representation, named product copula. More generally, hypergraph can also be formulated to be a special case of product copula with each variable dimension sub-copula corresponding to a sub-graph.

\section{Copula Entropy}
\label{s:CopEnt}
\subsection{Theory}
Copula theory is about representation of multivariate dependence with copula function \cite{joe1997,nelsen1998}. At the core of copula theory is Sklar theorem \cite{6} which states that multivariate probability density function can be represented as a product of its margins and copula density function which represents dependence structure among random variables. Such representation seperates dependence structure, i.e., copula function which contains all the dependence information, with margins of individual variables. Copula representation makes it possible to measure statistical dependence in copula function regardless of joint distribution and marginal distribution. This section is to define an statistical independence measure with copula. For clarity, please refer to \cite{ma2011} for notations.

With copula density, CE is define as follows \cite{ma2011}:
\begin{definition}[Copula Entropy]
	\label{d:ce}
	Let $\mathbf{X}$ be random variables with margins $\mathbf{u}$ and copula density $c(\mathbf{u})$. CE of $\mathbf{X}$ is defined as
	\begin{equation}
	H_c(\mathbf{X})=-\int_{\mathbf{u}}{c(\mathbf{u})\log{c(\mathbf{u})}}d\mathbf{u}.
	\end{equation}
\end{definition}

In information theory, MI and entropy are two different concepts \cite{12}. In \cite{ma2011}, Ma and Sun proved that they are essentially same -- MI is also a kind of entropy, negative CE, which is stated as follows: 
\begin{theorem}
	\label{thm1}
	MI of random variables is equivalent to negative CE:
	\begin{equation}
	I(\mathbf{X})=-H_c(\mathbf{X}).
	\end{equation}
\end{theorem}
\noindent
The proof of Theorem \ref{thm1} is simple \cite{ma2011}. There is also an instant corollary (Corollary \ref{c:ce}) on the relationship between information of joint probability density function, margins and copula density function.
\begin{corollary}
	\label{c:ce}
	\begin{equation}
	H(\mathbf{X})=\sum_{i}{H(X_i)}+H_c(\mathbf{X}).
	\end{equation}
\end{corollary}
The above results cast insight into the relationship between entropy, MI, and copula through CE, and therefore build a bridge between information theory and copula theory. CE itself provides a mathematical theory of statistical independence measure and is a perfect tool for estimating dependence structure.

\subsection{Estimation}
\label{s:est}
It has been widely considered that estimating MI is notoriously difficult. Under the blessing of Theorem \ref{thm1}, Ma and Sun  \cite{ma2011} proposed a simple and elegant non-parametric method for estimating CE (MI) from data which composes of only two steps:\\
$\bullet$ Step 1: Estimating Empirical Copula Density (ECD);\\
$\bullet$ Step 2: Estimating CE.\\

For Step 1, if given data samples $\{\mathbf{x}_1,\ldots,\mathbf{x}_T\}$ i.i.d. generated from random variables $\mathbf{X}=\{x_1,\ldots,x_N\}^T$, one can easily estimate ECD as follows:
\begin{equation}
F_i(x_i)=\frac{1}{T}\sum_{t=1}^{T}{\chi(\mathbf{x}_{t}^{i}\leq x_i)},
\end{equation}
where $i=1,\ldots,N$ and $\chi$ represents for indicator function. Let $\mathbf{u}=[F_1,\ldots,F_N]$, and then one can derives a new samples set $\{\mathbf{u}_1,\ldots,\mathbf{u}_T\}$ as data from ECD $c(\mathbf{u})$. In practice, Step 1 can be easily implemented non-parametrically with rank statistics.

Once ECD is estimated, Step 2 is essentially a problem of entropy estimation which has been contributed with many existing methods. Among them, the kNN method \cite{13} was suggested in \cite{ma2011}. With rank statistics and kNN methods, one can derive a non-parametric method of estimating CE, which can be applied to any situation without assumptions on the underlying system. 

Since the method for estimating CE is non-parametric, it can be applied to any cases for estimating dependence structure.

\section{Estimating Maximum Spanning Copula}

\subsection{The Framework based on Copula}
With copula, we propose a framework for estimating dependence structure regardless of properties of individual variables. In the framework, empirical copula is first estimated from data implicitly or explicitly, and then dependence structure is estimated from empirical copula based on dependence relationships which is measured with CE. The framework is distribution free since one do not have to make assumptions on and inference the parametric form of individual variables under the risk of model misspecification. The framework can be implemented non-parametrically since estimating CE involves only rank-based empirical copula and kNN entropy estimation. In this way, the framework focuses on only dependence structure and is inrelevant to properties of individual variables.

\subsection{Maximum Spanning Copula Problem}
Suppose we want to approximate dependence relations with a type of structure $\mathbf{T}(\mathbf{t})$, where $\mathbf{t}$ is the parameter specifying $\mathbf{T}$. Given a group of i.i.d. samples $X$ generated from a $N$ dimensional random vector $\mathbf{x}\in \mathcal{R}^N \sim p(\mathbf{x})$, an objective function $F$ can be defined on it, and be optimized to inference $\mathbf{T}$ with respect to $\mathbf{t}$:
\begin{equation}
\arg\max_{t}{F(t;X)}.
\end{equation}
Here, $t$ belongs to the structure associated with product copula, and $F$ is defined as the dependence information contained in $t$ so that such dependence information in $X$ are contained in $t$ as much as possible. 

Now consider a $N$ dimensional copula density $\mathbf{c}$ of $\mathbf{x}$. We can derive its empirical estimation $\mathbf{\hat{c}}$ from $X$, which contains all the dependence information in data. One of the natural idea is to build a $\mathbf{\hat{c}}$ covering the main dependence relationships with product copula (or graphical model). We call such product copula `\textit{maximum spanning copula}' (MSC). The MSC approximation of $\mathbf{c}$ in terms of bivariate product copula composes of a product of $N-1$ bivariate copula. Then the above $F$ can be defined by the sum of dependence measurements on $N-1$ subcopulas. The problem is how to find such a optimal product copula to approximate the dependence structure among random variables. 

\subsection{Construction Algorithm of MSC}
We propose a method for estimating from samples their MSC with Chow-Liu algorithm, which composes of two steps (Algorithm \ref{alg:estprodcop}). First, a dependence matrix is derived from data, each element of which is the value of CE between a pair of variables. Then based on dependence measure matrix, a structure $\mathbf{T}$ on $N$ random variables is built where the weight of each edge is equal to dependence between two variables. Constructing optimal product copula equals to finding MSC of $t$ from data. This is a well-defined problem, which can be solved by Chow-Liu algorithm \cite{11}. 

Chow-Liu algorithm here is actually an approach to construct Maximum Spanning Tree (MST) with CE as edge weights. There are some established MST algorithms, such as Kruskal's algorithm \cite{14} and Prim's algorithm \cite{15}.  Both algorithms can find the solution in polynomial time. We adopt Prim algorithm in our method (Algorithm \ref{alg:mst}). It starts with an empty edge set $E$, and then each time add from the complement set of $E$ one vertex $u$ and its corresponding edge $(u,v)$ with maximum weight so that $v\notin E$ has edge connection with $E$ and $(u,v)$ will has the maximum edge weight and make no loop in new $E$, till $E$ contains all the vertex.

\begin{algorithm}
	\caption{MSC estimation}
	\label{alg:estprodcop}
	\begin{algorithmic}
		\STATE {\bfseries Input:} data $\mathbf{x}$
		
		\STATE Calculate dependence matrix $\mathbf{C}_\mathbf{x}$ of $\mathbf{x}$ with CE estimation;
		\STATE Build dependence structure $\mathbf{T}$ with Algorithm \ref{alg:mst} based on $\mathbf{C}_\mathbf{x}$.
	\end{algorithmic}
\end{algorithm}

\begin{algorithm}
	\caption{MST algorithm}
	\label{alg:mst}
	\begin{algorithmic}
		\STATE {\bfseries Input:} edge set $E$, matrix of edge weight $W$
		
		\STATE $E_T = \{\}$;
		\REPEAT
		\STATE {Find the vertex $u$ which $W(u,v), v\in E\setminus E_T$ is with maximum weight;}
		\STATE {Add $u$ into $E_T$;}
		\UNTIL{$E_T = E$.}
		
	\end{algorithmic}
\end{algorithm}

%

\section{Experiments and Results}
\subsection{Simulated Data}
\begin{figure}
	\begin{center}
		\includegraphics[width=0.6\textwidth]{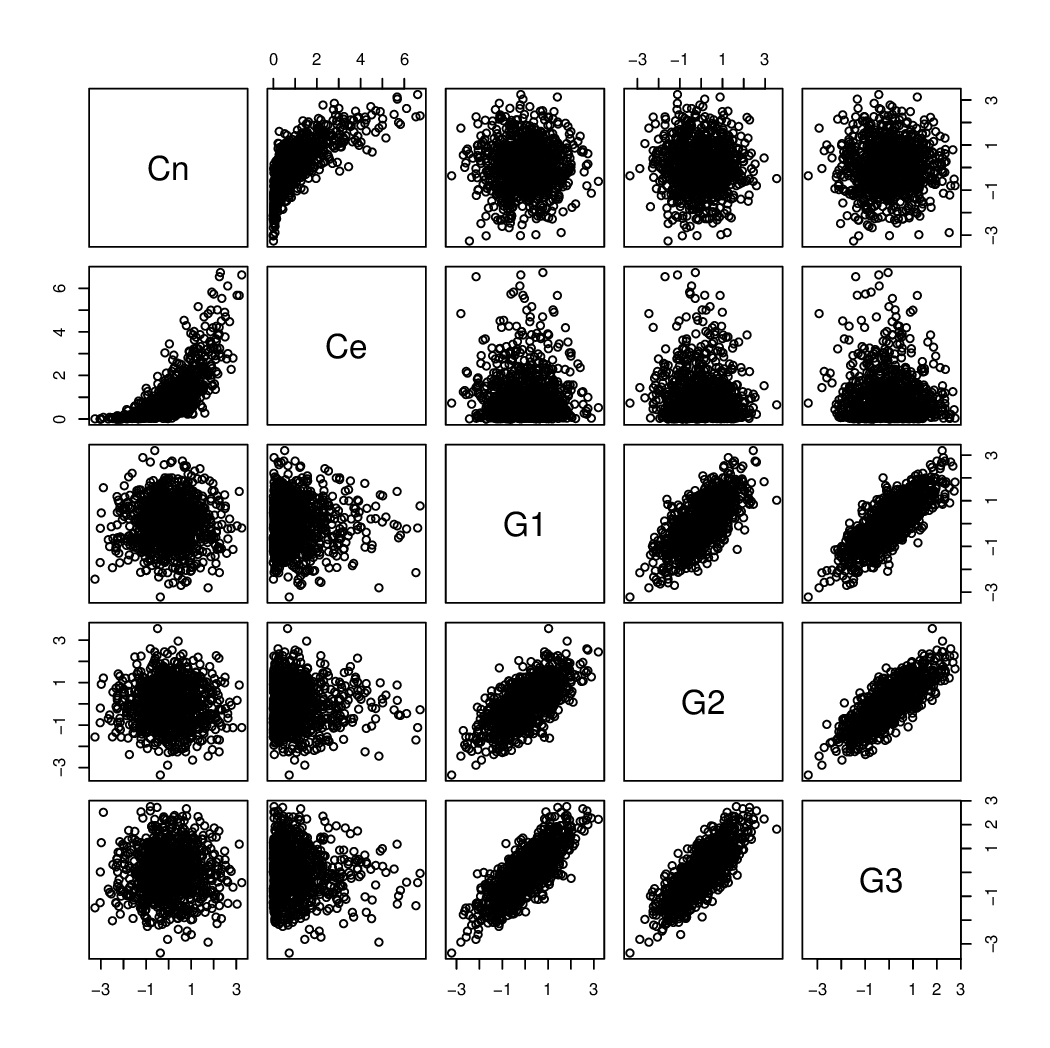}
		\caption{Simulated data.`G1-3' represent Gaussian variables, and `Cn',`Ce' represent two copula variable with normal and exponential margins.}
		\label{fig:data1}
	\end{center}
\end{figure}

We first test the proposed method on simulated data. A dataset with 1000 samples are randomly generated from 5 random variables, of which the first three ones are zero mean Gaussian and the others two are governed by Gaussian copula with margins as normal distribution and exponential distribution respectively (Figure \ref{fig:data1}). 

\begin{figure}
	\begin{center}
		\includegraphics[width=0.8\textwidth]{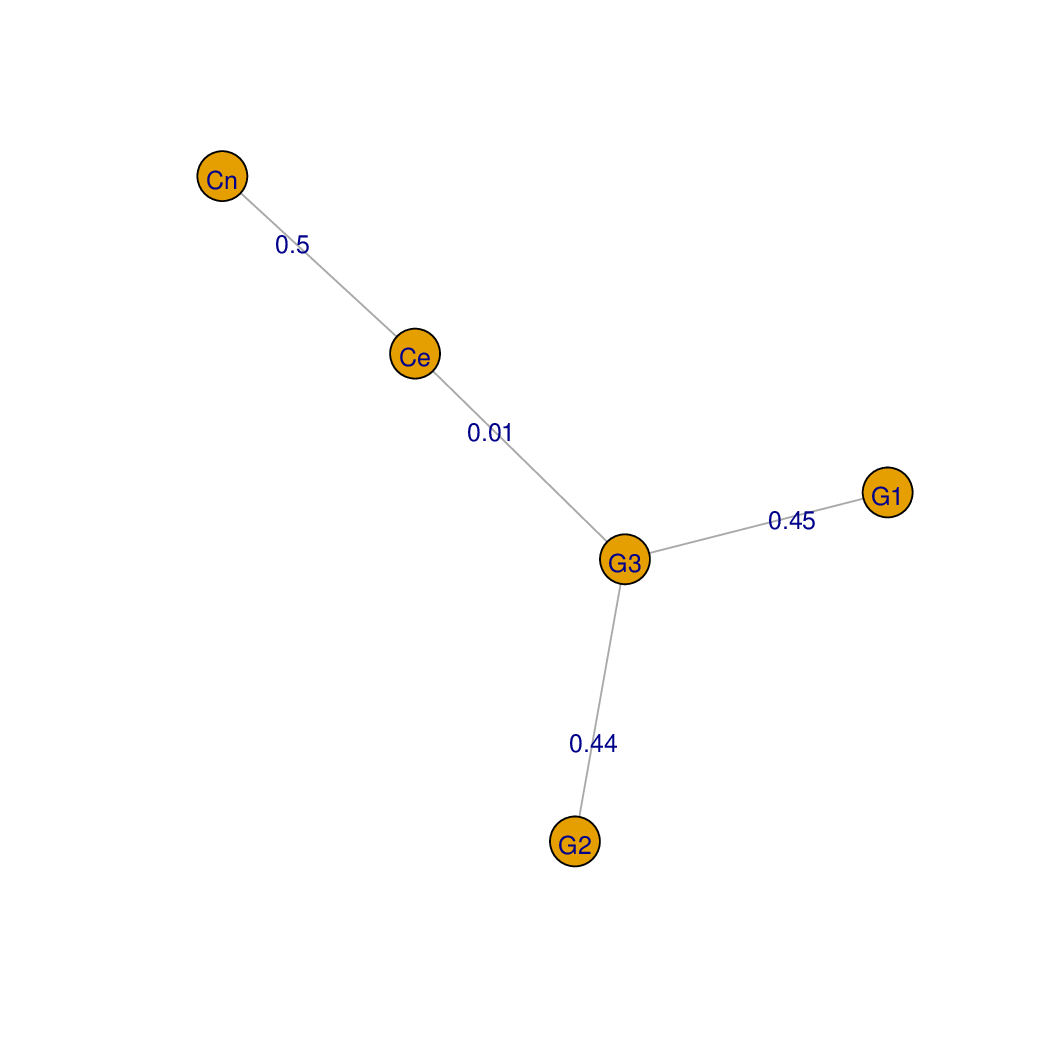}
		\caption{MSC of the simulated experiment.}
		\label{fig:exp1tree}
	\end{center}
\end{figure}

The algorithm \ref{alg:estprodcop} is run on the dataset and then a dependence tree is derived as illustrated in Figure \ref{fig:exp1tree}, which shows the dependence relationships of the two variable groups are correctly estimated.

\subsection{Real Datasets}
\subsubsection{Abalone Data}
Abalone dataset in UCI data repository \cite{16} was built to predict the age of abalone based on physical measurements of abalone body, such as weights and height. It composes of 4177 samples with 9 attributes. It was usually treated as a regression problem of predicting age with physical measurements. Here we focus on the dependence relations among the 9 attributes, which may benefit understanding this prediction task.

\begin{figure}
	\begin{center}
		\includegraphics[width=0.8\textwidth]{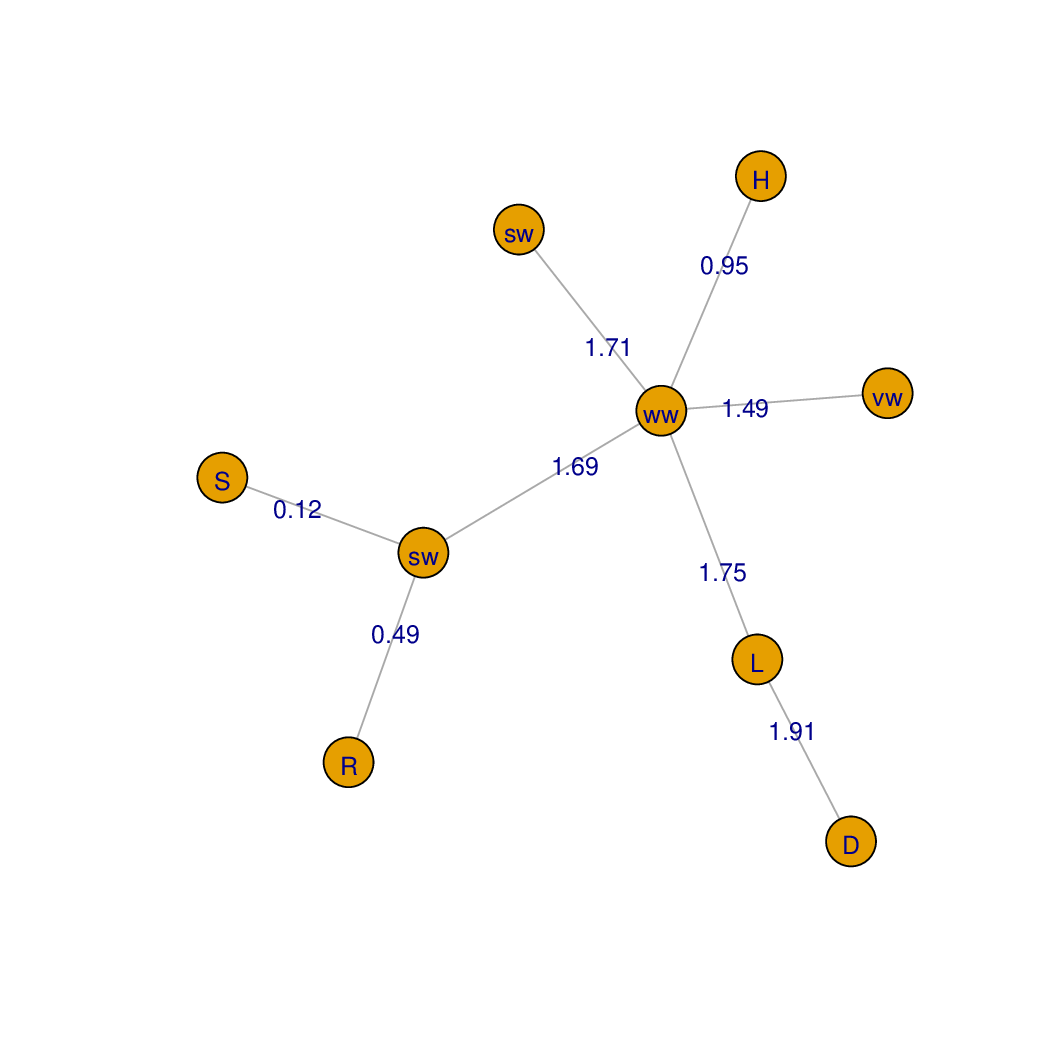}
		\caption{MSC of the Abalone dataset.}
		\label{fig:aba1mpt}
	\end{center}
\end{figure}

Figure \ref{fig:aba1mpt} shows one of all the MSC of abalone data in the experiments, where edges are labeled by CE weights. Except Sex and rings, other seven attributes are linked with relatively strong weighted edges. This dependence relationships between seven attributes can be interpreted as the reflection of abalone's body growth. It can also be learned that the edges between these seven attributes are the backbone of the estimated trees, while the nodes for `sex' and `ring' are weakly attached to other seven nodes. That implies that all the physical measurements increase as the abalones grow up, while ring and sex is not strongly related with these process. In this sense, we argue that predicting ring or age with the other attributes may not be reasonable.

\subsubsection{Housing Data}
The Boston house price dataset in UCI data repository \cite{16} is collected from a 1970 census, first published by Harrison, D. and Rubinfeld \cite{17}, with the aim to study how to predict `Medv' \footnote{For the abbr. of 14 attributes of Housing dataset, please refer to UCI machine learning dataset websit \cite{16}.} from the other 13 attributes. It includes 506 samples with 14 mixed type attributes, including 13 continuous attributes and 1 binary one. Previous researches mainly treat it as a regression problem without considering the relationships between attributes. In this experiment, we try to understand the data by studing the dependence structure.

\begin{figure}
	\begin{center}
		\includegraphics[width=0.8\textwidth]{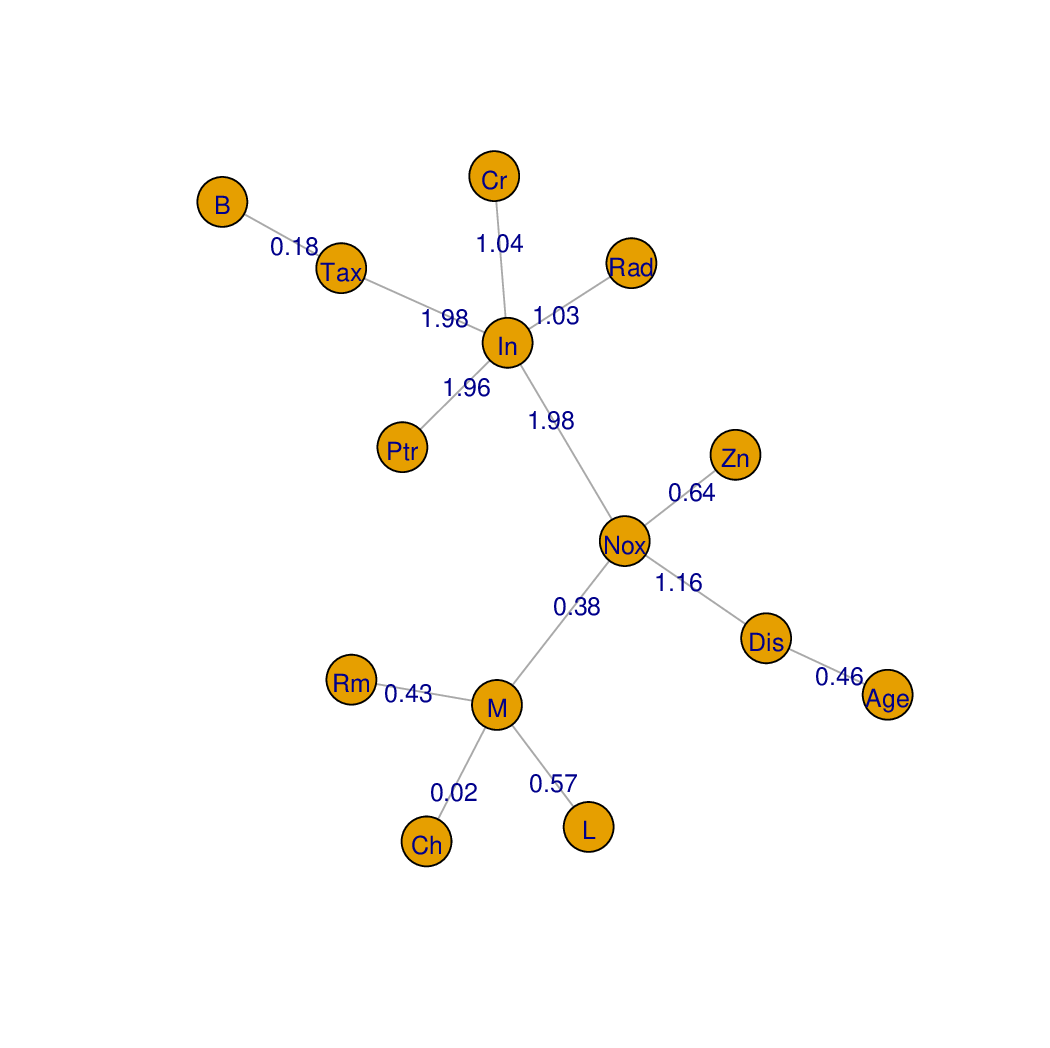}	\caption{MSC of the Housing dataset.}
		\label{fig:housingmitree}
	\end{center}
\end{figure}

The MSC algorithm was run on the housing dataset. A dependence tree was generated, which is plotted in Figure \ref{fig:housingmitree}. Experimental results showed that two groups of linked edges including `Crim-Indus-Rad-Tax-Ptratio-Nox-Dis' and `Medv-Lstat', remain strong in the estimated tree. So predicting `Medv' from all other attributes is problematic.

\section{Discussion}
Our philosophy of structure learning is that the more we know, the better structure we can learn. Using copula, one can incorporate all the dependence information without model constraints and meanwhile all the information is about nothing but dependence relations. Then structure learning based on copula provides a general framework which can unify all the related structure learning methods. Another main advantage is such estimation is irrelevant to properties of individual variables. 

\begin{figure}
	\begin{center}
		\includegraphics[width=0.6\textwidth]{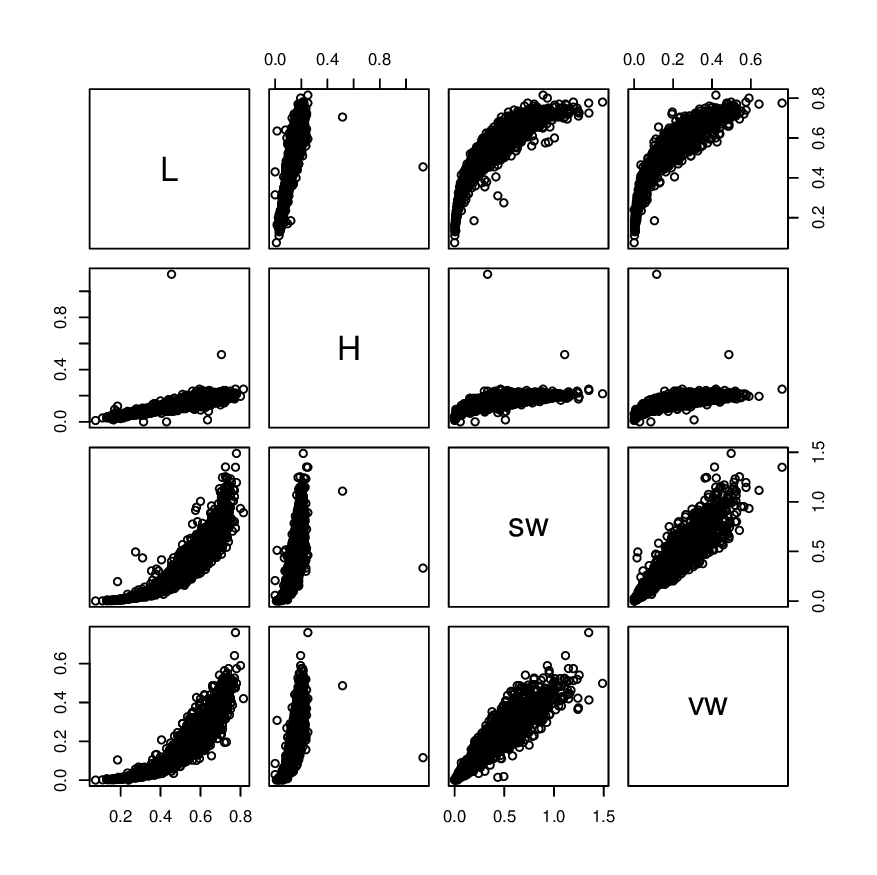}
		\caption{Scatter plot of four attributes in Abalone dataset.`L',`H',`sw',`vw' represent Length, Height, Shucked weight, and Viscera weight.}
		\label{fig:aba1}
	\end{center}
\end{figure}

\begin{figure}
	\begin{center}
		\includegraphics[width=0.6\textwidth]{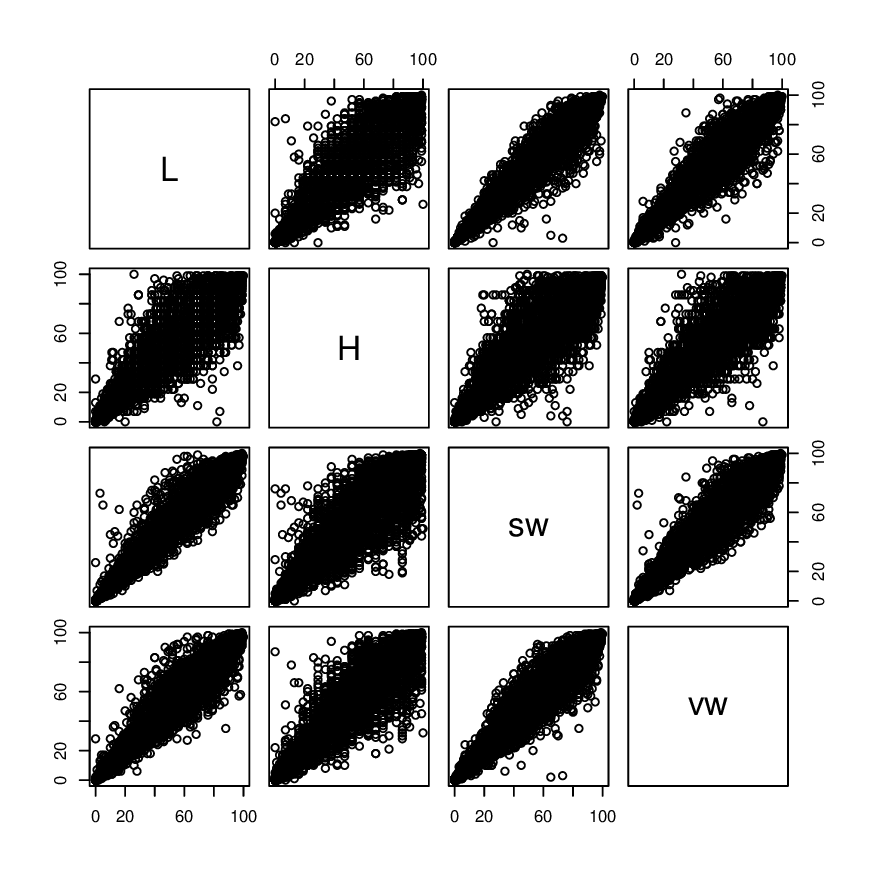}
		\caption{Empirical copula density of four attributes in Abalone dataset.}
		\label{fig:cpaba1}
	\end{center}
\end{figure}

Due to copula, MSC estimation has some advantages. The first one is insensitive to outlier. For instance, there are a few outliers in the abalone dataset. Traditionally, they should be eliminated otherwise they may cause large deviation in the following dependence measures calculation. But here it is unnecessary because estimation of CE is less susceptible to outliers. Comparing the original data (Figure \ref{fig:aba1}) with its copula (Figure \ref{fig:cpaba1}), one can learn that outliers cause almost no effect on empirical copula and hence the estimation of MSC. 

Besides robustness to outliers, we highlight another advantage made possible by copula, i.e., the ability to deal with non-Gaussianity within data. It can be observed from Figure \ref{fig:aba1} that all the attributes possesses non-gaussianity to some extent which is demonstrated in their joint densities with other attributes. While all the pairwise copulas show a very similar dependency structure after marginal properties of variables are separated from joint distribution as illustrated in Figure \ref{fig:cpaba1}. 

Copula also makes non-linear transformation on data unnecessary. When learning structure, some researchers proposed to transform the data into a suitable scale before further dependence analysis, through monotonically increasing function, such as normalization, nonlinear exponential/log functions. In our experiment, it is unnecessary due to copulas invariant to such kind of transformation.

Dependence representation using only bivariate dependence is limited. Given $N(N-1)$ pair dependence relations of $N$ random variable, only $N-1$ of relations compose of tree structure. However, MSC can show the skeleton of dependence structure within data, which reflect the main relationships. It would help to understand the data better at first.

\section{Conclusion and Further Direction}
In the paper, we propose a framework for estimating dependence structure using copula. Copula can represent all kinds of dependence relations among random variables, and makes no additional assumption on the underlying distributions. Graphical models is considered as a special case in the copula framework, named product copula. In this framework, we proposed a method for estimating MSC with a Chow-Liu like method based on CE. The proposed method was demonstrated on simulated data and two real dataset to estimate their dependence structure. Experimental results showed that the estimated MSC can benefit us understanding the data better and that the copula framework can estimate dependence representations which are margin-free, robust to outlier, and invariant to increasing transformation. 

Theory of copula and CE are about representation and measurement of statistical dependence respectively. They provide a sound theoretical base for further research. The problem of structure learning is to infer the underlying dependence relationship. For structure learning with copula, many problems remain to be tackled. For example, how to choose different structures (or family when in a parametric way) for applications? MSC represents the simple dependence relations. In the further, we will study the estimation of more complex dependence structures through copula. This can imply many real applications in biological, social, and physical sciences.



\end{document}